\documentclass[letterpaper]{article} % DO NOT CHANGE THIS
\usepackage{arxiv24}  % DO NOT CHANGE THIS
\usepackage{times}  % DO NOT CHANGE THIS
\usepackage{helvet}  % DO NOT CHANGE THIS
\usepackage{courier}  % DO NOT CHANGE THIS
\usepackage[hyphens]{url}  % DO NOT CHANGE THIS
\usepackage{graphicx} % DO NOT CHANGE THIS
\usepackage{natbib}  % DO NOT CHANGE THIS AND DO NOT ADD ANY OPTIONS TO IT
\usepackage{caption} % DO NOT CHANGE THIS AND DO NOT ADD ANY OPTIONS TO IT
\usepackage{algorithm}
\usepackage{algorithmic}
\usepackage{subcaption}
\usepackage{listings}
\usepackage{color}

\usepackage{amsmath}
\usepackage{amsfonts}
\usepackage{multicol}
\usepackage{multirow}
\usepackage{dirtytalk}

\usepackage{newfloat}
\usepackage{listings}
\DeclareCaptionStyle{ruled}{labelfont=normalfont,labelsep=colon,strut=off} % DO NOT CHANGE THIS
\floatstyle{ruled}
\newfloat{listing}{tb}{lst}{}
\floatname{listing}{Listing}
\title{How to Mask in Error Correction Code Transformer: Systematic and Double Masking}
\author{Seong-Joon Park\textsuperscript{\rm 1},
    Hee-Youl Kwak\textsuperscript{\rm 2},
    Sang-Hyo Kim\textsuperscript{\rm 3},
    Sunghwan Kim\textsuperscript{\rm 2},
    Yongjune Kim\textsuperscript{\rm 4}\thanks{Corresponding author.},
    Jong-Seon No\textsuperscript{\rm 1}
}
\affiliations{
    \textsuperscript{\rm 1}Department of Electrical and Computer Engineering, Seoul National University\\
    \textsuperscript{\rm 2}School of Electrical Engineering, University of Ulsan\\
    \textsuperscript{\rm 3}Department of Electrical and Computer Engineering, Sungkyunkwan University\\
    \textsuperscript{\rm 4}Department of Electrical Engineering, Pohang University of Science and Technology (POSTECH)\\
    \{joon2247, jsno\}@snu.ac.kr, \{hykwak, sungkim\}@ulsan.ac.kr, iamshkim@skku.edu, yongjune@postech.ac.kr
}

\usepackage{bibentry}
\begin{document}

\maketitle

\begin{abstract}
In communication and storage systems, error correction codes~(ECCs) are pivotal in ensuring data reliability.
As deep learning's applicability has broadened across diverse domains, there is a growing research focus on neural network-based decoders that outperform traditional decoding algorithms.
Among these neural decoders, Error Correction Code Transformer~(ECCT) has achieved the state-of-the-art performance among neural network-based decoders, outperforming other methods by large margins.
To further enhance the performance of ECCT, we propose two novel methods.
First, leveraging the systematic encoding technique of ECCs, we introduce a new masking matrix for ECCT, aiming to improve the performance and reduce the computational complexity.
Second, we propose a novel transformer architecture of ECCT called a double-masked ECCT.
This architecture employs two different mask matrices in a parallel manner to learn more diverse features of the relationship between codeword bits in the masked self-attention blocks.
Extensive simulation results show that the proposed double-masked ECCT outperforms the conventional ECCT, achieving the state-of-the-art decoding performance among neural network-based decoders with significant margins.

\end{abstract}

\section{Introduction}

Over recent years, deep learning methods have experienced rapid advancements and achieved phenomenal success in various tasks, such as natural language processing (NLP), image classification, object detection, semantic segmentation, etc.
Among the various deep learning architectures available, the transformer architecture~\cite{b_transformer} has consistently demonstrated the state-of-the-art results in most tasks.
After breaking through in NLP, the application of transformer has expanded to include computer vision tasks~\cite{b_vit,b_swin}, and again demonstrated outstanding performances compared to the conventional neural network architecture.
The versatility of transformer has now extended to the field of error correction codes (ECCs)~\cite{b_ECCT}. 

ECCs have played a pivotal role in ensuring reliability in wireless communication and storage systems by serving as a key technology to correct errors in noisy environments.
The ECC research based on deep learning architectures has primarily focused on decoders that offer superior error correction performance.
Stimulated by the advancement of deep learning techniques, a new type of decoders based on neural networks has emerged~\cite{b_comm1,b_comm2,b_comm3,b_comm4}.
These neural network-based decoders have overcome the limitations of traditional algorithm-based decoders.
Notably, among them, the transformer-based ECC decoder achieves the state-of-the-art performance.

The transformer based ECC decoder, also called Error correction Code Transformer (ECCT)~\cite{b_ECCT}, applied a mask matrix in the self-attention block to learn the noise in the communication channel.
Since not all bits in a codeword are equally related, ECCT can improve the performance by using a mask matrix that facilitates learning of the relevance between codeword bits.
In the conventional ECCT work, the mask matrix is derived from the parity check matrix (PCM) whose parity check equations determine a direct relationship between codeword bits.

However, the problem is that numerous PCMs exist for the same codebook, which raises the following question:\\
\say{Which one of those PCMs is optimal to aid the self-attention mechanism in ECCT?}\\
We find that selecting a different PCM~(i.e., different mask matrix) has a crucial impact on the performance of ECCT.
Hence, identifying the optimal PCM for constructing a mask matrix is an important problem of ECCT, which has not investigated yet.

In this work, we first introduce a novel mask matrix specifically tailored for ECCT, which we term a \emph{systematic mask matrix} due to its construction based on a systematic PCM.
The systematic mask matrix has more masking positions than the mask matrix used in the conventional ECCT.
This property of the systematic mask matrix makes the self-attention map \emph{sparser}, and enables more compact and concentrated learning of the relationship between codeword bits.
{It is a surprising observation given that systematic form of the matrix is typically utilized for efficient encoding rather than decoding~\cite{b_modern}.}
Yet, in ECCT, they play a pivotal role in enhancing decoding performance, which is an interesting finding.

Next, we propose a novel transformer architecture called a \textit{double-masked~{\rm(DM)} ECCT}.
This architecture consists of two parallel masked self-attention blocks, each employing a distinct mask matrix.
Utilizing two different mask matrices enables the DM ECCT to capture diverse features of the bit relationship.

We apply the proposed methods to two representative ECCs: Bose–Chaudhuri–Hocquenghem~(BCH) and polar codes as in~\cite{b_ECCT}.
Through extensive simulations across diverse code parameters, we demonstrate that \emph{systematic mask matrices} improve decoding performance, while reducing decoding complexity.
Furthermore, our DM ECCT, integrating both systematic and conventional mask matrices, achieves the state-of-the-art performance among neural network-based decoders for both BCH and polar codes.
To the best of our knowledge, this is the first work to propose a new mask matrix structure tailored for ECCT.
Additionally, our proposed DM ECCT is a novel transformer architecture for ECC to utilize multiple mask matrices, adding diversity to the ECCT architecture.

\section{Related Works}
In this section, we briefly review the deep learning approaches to ECC applications, such as the neural network-based ECC decoders.
There are mainly two approaches: The model-based approach and the model-free approach.
\subsection{Model-Based Approach}
The first approach is to implement a conventional decoding methods (e.g., belief propagation~(BP) decoder and min-sum~(MS) decoder) on a neural network.
These neural decoders unfold the iterative decoding operation on the Tanner graph into a deep neural network.
Nachmani et al.~\cite{b_1} proposed a neural decoder based on the recurrent neural network for BCH codes and achieved the performance improvement over the standard BP decoder.
Dai et al.~\cite{b_2} modified the neural MS decoder for protograph low-density parity-check~(LDPC) codes. A parameter sharing mechanism was proposed for training scalability to long codes, which also reduces the training complexity and memory cost. 
Furthermore, a number of studies exhibited that neural network-based BP and MS decoders outperform the traditional decoders~\cite{b_3, b_4, b_5, b_6, b_7}.
However, these model-based neural decoders might face restrictive performance limits due to architectural constraints.

\subsection{Model-Free Approach}
The second approach is a model-free approach, which employs neural network architectures with no prior knowledge of decoding algorithms.
This approach is not restricted to the conventional decoding models but encounters a significant challenge initially.
The model-free approach faces the overfitting problem since it is impractical to train all codewords in a codebook.
However, Bennatan et al.~\cite{b_preproc} proposed a preprocessing that enables the black box model decoder to overcome the overfitting problem.
They utilized the syndrome to learn noise only with the all-zero codeword.
Also, they combined recurrent neural network architecture with the preprocessing.
ECCT is another work that implemented the transformer without the overfitting problem using the same preprocessing and achieved excellent decoding performance through the masked self-attention mechanism.
However, in the previous ECCT research, the mask matrix was directly derived from the PCM of the conventional decoding algorithm, rather than being adjusted for ECCT.

%Kwak et al. applied a neural MS decoder for generalized low-density parity-check~(LDPC) codes and showed the proposed decoder outperform the conventional MS decoder for generalized LDPC codes.

\section{Background}

In this section, we briefly summarize some background on the ECCs and the preprocessing and postprocessing of the conventional ECCT~\cite{b_ECCT}.

\subsection{Error Correction Codes}
Let $C$ be a linear code.
A codeword $x\in C \subset \{0,1\}^n$ can be defined by a generator matrix $G$ of size $k\times n$ and a PCM $H$ of size $(n-k)\times n$, which satisfies $GH^T=0$ over $\{0,1\}$ with modulo $2$ addition.
In other words, a codeword $x$ can be determined by the constraint $Hx=0$. 
Let $x_s$ be a binary phase shift keying modulation of $x$~(i.e., $x_s=+1$ if $x=0$ and $x_s=-1$ if $x=1$) and let $y$ be a channel output of $x_s$ after passing the additive white Gaussian noise channel ($y=x_s+z$, where $z$ denotes Gaussian random noise, i.e., $z\sim N(0,\sigma^2)$).

The objective of the decoder~($f:\mathbb{R}^n \rightarrow \mathbb{R}^n$) is to recover the original transmitted codeword $x$ by correcting errors.
When $y$ is received, the decoder first determines if the received signal is corrupted by checking the syndrome $s(y)=Hy_b$, where $y_b=\text{bin}(\text{sign}(y))$. Here, $\text{sign}(a)$ represents $+1$ if $a\ge 0$ and $-1$ otherwise and $\text{bin}(-1)=1$, $\text{bin}(+1)=0$.
If $s(y)$ is non-zero vector, it is detected that $y$ is distorted during the transmission, and the decoder initiates the error correction process.
ECCT~\cite{b_ECCT} employs the transformer architecture to approximate the role of the ECC decoder.

\subsection{Preprocessing and Postprocessing}

By employing the preprocessing method in~\cite{b_preproc}, the all-zero codeword is enough for training, which makes the ECCT robust to overfitting problems.
The preprocessing conducts input embedding as $\tilde{y} = [|y|,s(y)]$, where $|y|$ is the magnitude of $y$, and $[\cdot,\cdot]$ denotes the concatenation of two vectors.
The objective of the ECCT is to estimate the multiplicative noise $\tilde{z}_s$, which is defined by
\begin{equation}
\label{equ_multi_noise}
    y=x_s+z = x_s\tilde{z}_s.
\end{equation}
Since the ECCT estimates the multiplicative noise, \mbox{$f(y)=\hat{z}_s$} and the estimation of $x$ is
$\hat{x} = \text{bin}(\text{sign}(yf(y)))$.
If the multiplicative noise is correctly estimated, then \mbox{$\text{sign}(\tilde{z}_s)=\text{sign}(\hat{z}_s)$} and $\text{sign}(\tilde{z_s}\hat{z}_s)=1$.
In this case, $\hat{x}$ is obtained by

\begin{align*}
    \hat{x} &= \text{bin}(\text{sign}(yf(y)))= \text{bin}(\text{sign}(x_s\tilde{z_s}\hat{{z}_s}))\\
    &=\text{bin}(\text{sign}(x_s))\\
    &=x.
\end{align*}

\begin{figure}[t!]
\centering
\begin{subfigure}[b]{\columnwidth}
    \centering
    \includegraphics[width=\columnwidth]{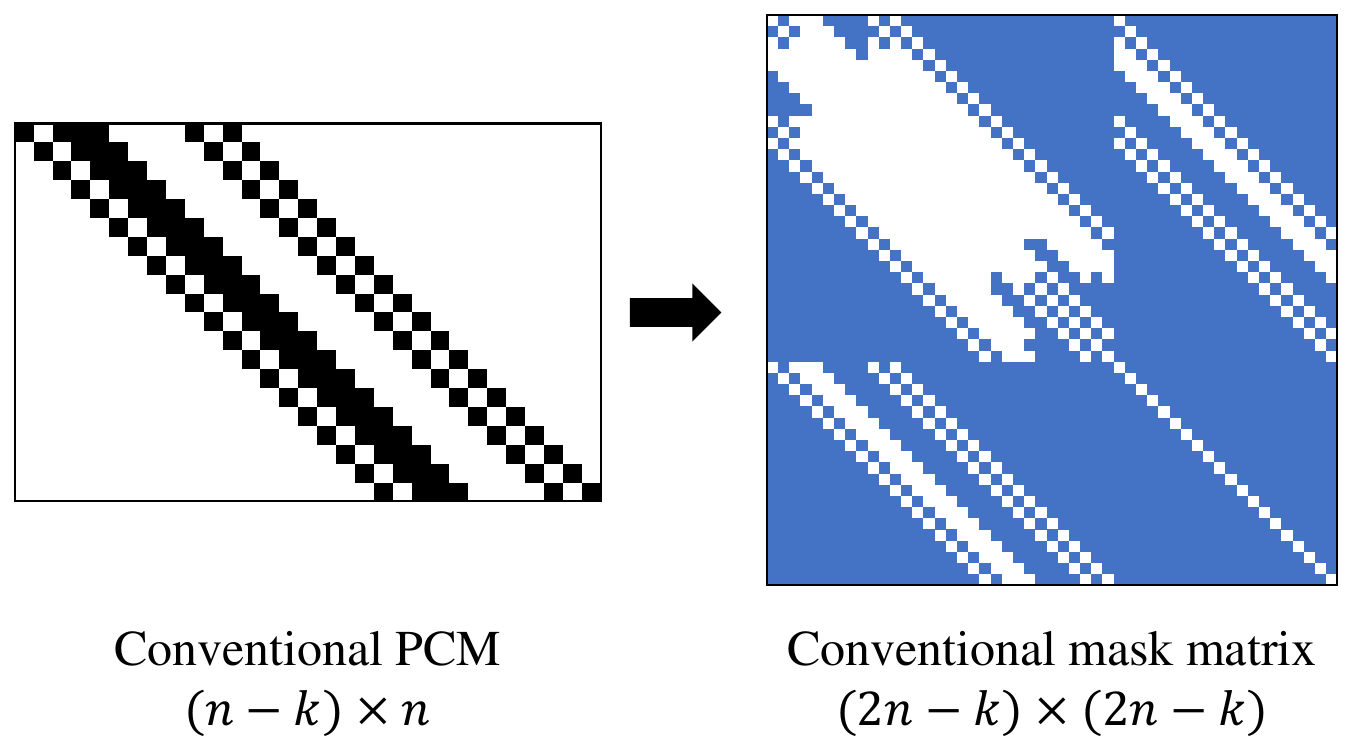} % Reduce the figure size so that it is slightly narrower than the column. Don't use precise values for figure width.This setup will avoid overfull boxes.
\caption{Conventional mask matrix\label{subfig_non}}

\end{subfigure}
\hfill
\begin{subfigure}[b]{\columnwidth}
    \centering
    \includegraphics[width=\columnwidth]{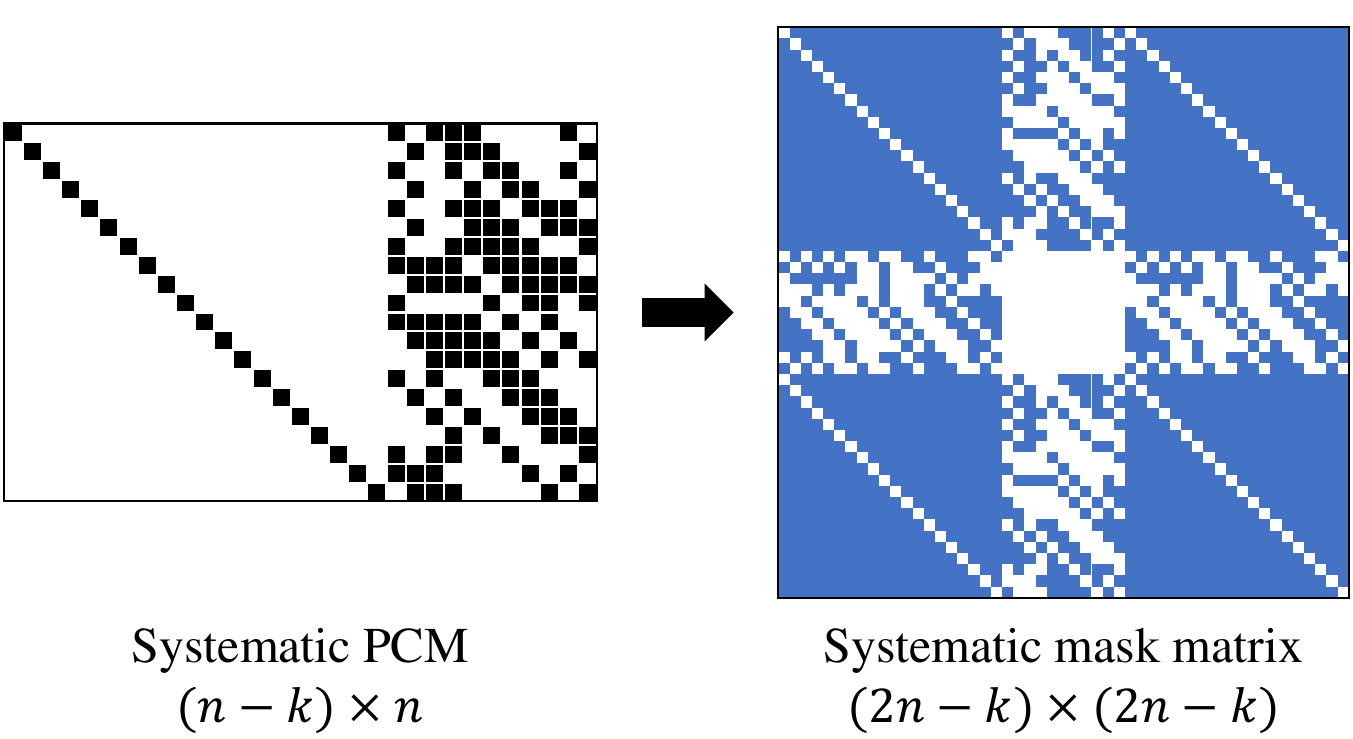} % Reduce the figure size so that it is slightly narrower than the column. Don't use precise values for figure width.This setup will avoid overfull boxes.
\caption{Systematic mask matrix\label{subfig_sys}}

\end{subfigure}
\caption{Two different types of mask matrices: (a) The conventional mask is determined by the conventional PCM~\cite{b_ECCT}, (b) the systematic mask matrix is determined by the systematically transformed PCM.
The non-zero entries in PCMs and masking positions in the mask matrix are depicted in colored boxes.
The proposed systematic mask matrix masks more values than the conventional mask matrix, resulting in sparser self-attention maps.
\label{fig_mask}}
\end{figure}

\section{Proposed Methods}

\begin{figure}[t!]
\centering
\includegraphics[width=\columnwidth]{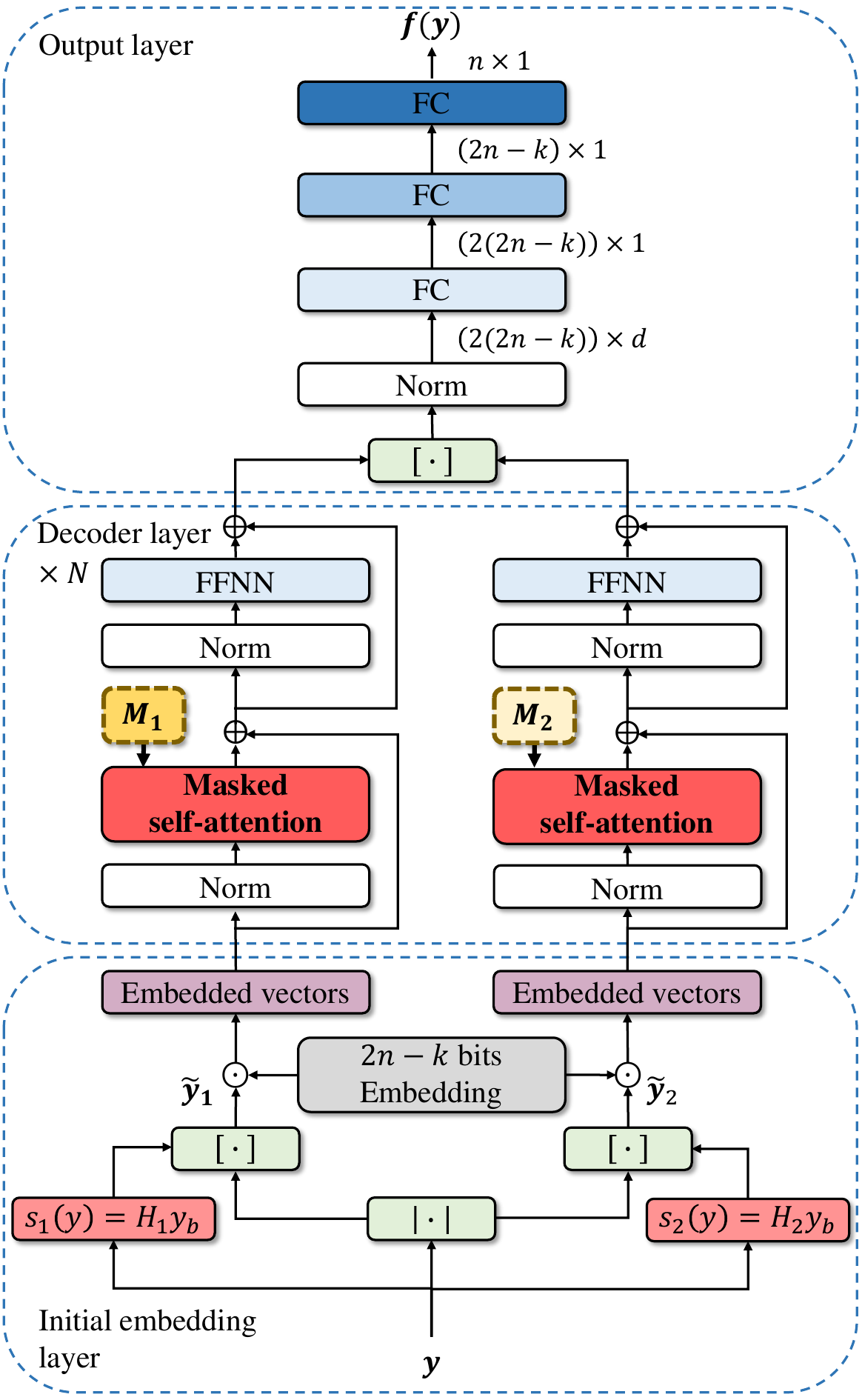}
\caption{Architecture of the DM ECCT.
\label{fig_dec_arch}}
\end{figure}

We propose the systematic mask matrix for the ECCT architecture and the DM ECCT with two different mask matrices.
The proposed systematic mask matrix and DM ECCT architecture are shown in Figures~\ref{fig_mask} and \ref{fig_dec_arch}, respectively.

\subsection{Systematic Mask}

In the ECCT, the mask matrix constructed by the PCM $H$ facilitates the learning.
Since the mask matrix is uniquely determined by the PCM.
There is one-to-one correspondence between the PCM and the mask matrix.
Unlike the ECCT in~\cite{b_ECCT}, we construct the systematic mask matrix from a specific PCM, which is defined as systematic PCM. 
{To construct the systematic mask matrix, we first transform the PCM as a systematic form~(i.e., reduced row echelon form) by Gaussian elimination technique.}
The systematically formed PCM $H_{\rm sys}$ is expressed as $H_{\rm sys} = [I_{n-k}~P]$, where $I_{n-k}$ is the identity matrix of size $(n-k)\times(n-k)$ and $P$ is a matrix of size $(n-k)\times k$.
{Then, we can construct the systematic mask matrix from $H_{\rm sys}$ by Algorithm~\ref{alg_mask}, which is slightly modified from~\cite{b_ECCT}.}
The conventional mask matrix~\cite{b_ECCT} and the proposed systematic mask matrix for the BCH code $(31, 11)$ are compared in Figure~\ref{fig_mask}.

%def row_reduce(PCM):
%    assert PCM.ndim == 2
%    nrows, ncols = PCM.shape
%    sys_PCM = PCM.copy()
%    pivot = 0
%    for j in range(ncols):
%        idxs = pivot + np.nonzero(sys_PCM[pivot:,j])[0]
%        if idxs.size == 0:
%            continue
%        sys_PCM[[pivot,idxs[0]],:] = %sys_PCM[[idxs[0],pivot],:]
%        idxs = np.nonzero(sys_PCM[:,j])[0].tolist()
%        idxs.remove(pivot)
%        sys_PCM[idxs,:] = (sys_PCM[idxs,:] + %sys_PCM[pivot,:]) % 2
%        pivot += 1
%        if pivot == sys_PCM.shape[0]:
%            break
%    return sys_PCM

\begin{algorithm}[!t]
\caption{Mask matrix construction}

\label{alg_mask}
\begin{lstlisting}[mathescape=true]
def g(systematic, conv_PCM):
    if systematic is True:
        PCM = make_sys(conv_PCM)
    else:
        PCM = conv_PCM
    nrow, ncol = PCM.size
    mask_size = ncol + nrow
    mask = eye(mask_size, mask_size)
    for ii in range(nrow):
        idx = where(PCM[ii] > 0)[0]
        for jj in idx:
            for kk in idx:
                mask[jj, kk] += 1
                mask[kk, jj] += 1
                mask[N + ii, jj] += 1
                mask[jj, N + ii] += 1
    return $-\infty$ ($\neg$ mask) 
\end{lstlisting}
\end{algorithm}

As shown in Figure~\ref{fig_mask}(b), the first $(n-k)\times (n-k)$ submatrix of the systematic mask matrix is the identity matrix $I_{n-k}$ of $H_{\rm sys}$, which is for efficient encoding.
Also, the systematic mask matrix has more masking positions compared to the conventional mask matrix.
In other words, employing the systematic mask matrix makes the masked self-attention map sparser than the conventional mask matrix.
As the masked self-attention map becomes sparse, the model tends to focus more on the unmasked locations.
Since not all positions in the codeword are equally related, it is important to focus on the highly related positions in the codeword during the self-attention mechanism.
Therefore, employing an ``appropriate" mask matrix leads to better training and decoding for ECCT.
Furthermore, since only unmasking positions in the self-attention map participate in the training, the mask matrix with a large portion of masking positions leads to a lower training complexity.
Due to these reasons, the proposed ECCT using a systematic mask matrix, which we call a systematic-masked~(SM) ECCT from now on, is expected to have a better \mbox{complexity-decoding performance tradeoff} than the conventional ECCT.

We propose to use the systematic mask matrix for the masked self-attention mechanism instead of the conventional mask matrix.
Changing the conventional mask matrix to the systematic mask matrix leads to the enhancement in both error correction performance and computational complexity of ECCT.
This improvement is not limited to specific code lengths or code rates but to a wide range of coding parameters for various ECCs.

\subsection{Double-Masked ECCT}

To further improve the decoding performance, we propose a novel architecture called a DM ECCT.
In the DM ECCT, we utilize the two different mask matrices for two input embedded vectors in the masked self-attention module.
The overall architecture of the DM ECCT is illustrated in Figure~\ref{fig_dec_arch}.

During the initial embedding of the DM ECCT, the received codeword $y$ is converted to $\tilde{y}_1 = [|y|,s_1(y)]$ and $\tilde{y}_2 = [|y|,s_2(y)]$, where $s_1(y)=H_1y_b$ and $s_2(y)=H_2y_b$.
{$H_1$ and $H_2$ can be any PCMs that have the same codebook.}
Then, $\tilde{y}_1$ and $\tilde{y}_2$ with $2n-k$ bits are projected to $d$ dimensional embedding.
After passing the initial embedding layer, the decoder is defined as a concatenation of $N$ decoder layers.
The decoder layer consists of one masked self-attention, followed by one feed-forward neural network~(FFNN).
For each step, a normalization layer precedes and a residual connection is established.
Finally, the output layer takes the concatenated vector of two output vectors of the decoder layer.
It consists of three fully connected~(FC) layers, applied after a normalization layer.
The first FC layer reduces $2\times(2n-k)$ dimension vector to $2n-k$ dimension, the second reduces $d$ dimensional embedding to a one-dimensional $2n-k$ vector, and the third reduces $2n-k$ to an $n$ dimensional vector.
The output represents an estimate of soft multiplicative noise \mbox{$f(y)=\hat{z}_s$} with which the decoding is completed by bit flipping.

A key structure of a DM ECCT is that we employ two different mask matrices for two input vectors.
The input vector determined by $\tilde{y_i}$ is masked with $M_i$, since both $\tilde{y_i}$ and $M_i$ are determined by $H_i$, for $i=1,2$.
As mentioned above, the relevance between codeword bits are associated in the PCM, and numerous PCMs exist for the same codebook.
Rather than utilizing a single PCM (or a single mask matrix), utilizing a pair of PCMs can provide a diversity to the decoder.
In the decoder layer, the proposed DM ECCT architecture captures distinct features generated by two different PCMs.
These features are subsequently fused using concatenation and processed by the FC layers in the output layer.
Such a design allows the DM ECCT to adeptly discern the relationship between codeword bits, ultimately achieving state-of-the-art decoding performance among neural network-based decoders.

\begin{table*}[!t]
\centering
\begin{tabular}{ccccccccc}
\hline\hline
\multicolumn{9}{c}{$N=2$}                                                                                                                                                                                                                                                                                                                                                                                                \\ \hline\hline
\multicolumn{2}{c|}{$d$}                                                                                                    & \multicolumn{2}{c|}{$32$}                                                            & \multicolumn{2}{c|}{$64$}                                                            & \multicolumn{3}{c}{$128$}                                                                                          \\ \hline
\multicolumn{1}{c|}{Code}                                                                     & \multicolumn{1}{l|}{SNR}  & ECCT      & \multicolumn{1}{c|}{\begin{tabular}[c]{@{}c@{}}SM\\ ECCT\end{tabular}} & ECCT      & \multicolumn{1}{c|}{\begin{tabular}[c]{@{}c@{}}SM\\ ECCT\end{tabular}} & ECCT     & \begin{tabular}[c]{@{}c@{}}SM\\ ECCT\end{tabular} & \begin{tabular}[c]{@{}c@{}}DM\\ ECCT\end{tabular} \\ \hline\hline
\multicolumn{1}{c|}{\multirow{3}{*}{\begin{tabular}[c]{@{}c@{}}BCH\\ $(31,11)$\end{tabular}}}   & \multicolumn{1}{c|}{$4$~dB} & $3.43e-2$ & \multicolumn{1}{c|}{$1.77e-2$}                                         & $2.74e-2$ & \multicolumn{1}{c|}{$1.17e-2$}                                          & $1.97e-2$ & $9.06e-3$                                          & $\mathbf{6.11e-3}$                                \\ \cline{2-2}
\multicolumn{1}{c|}{}                                                                         & \multicolumn{1}{c|}{$5$~dB} & $1.47e-2$ & \multicolumn{1}{c|}{$6.12e-3$}                                         & $1.06e-2$ & \multicolumn{1}{c|}{$3.46e-3$}                                          & $6.60e-3$ & $2.36e-3$                                          & $\mathbf{1.38e-3}$                                \\ \cline{2-2}
\multicolumn{1}{c|}{}                                                                         & \multicolumn{1}{c|}{$6$~dB} & $5.02e-3$ & \multicolumn{1}{c|}{$1.70e-3$}                                         & $3.05e-3$ & \multicolumn{1}{c|}{$7.12e-4$}                                          & $1.66e-3$ & $4.43e-4$                                          & $\mathbf{1.78e-4}$                                \\ \hline
\multicolumn{1}{c|}{\multirow{3}{*}{\begin{tabular}[c]{@{}c@{}}BCH\\ $(31,16)$\end{tabular}}}   & \multicolumn{1}{c|}{$4$~dB} & $1.46e-2$ & \multicolumn{1}{c|}{$1.06e-2$}                                         & $1.20e-2$ & \multicolumn{1}{c|}{$8.22e-3$}                                          & $8.38e-3$ & $5.63e-3$                                          & $\mathbf{4.07e-3}$                                \\ \cline{2-2}
\multicolumn{1}{c|}{}                                                                         & \multicolumn{1}{c|}{$5$~dB} & $4.40e-3$ & \multicolumn{1}{c|}{$3.05e-3$}                                         & $3.39e-3$ & \multicolumn{1}{c|}{$2.09e-3$}                                          & $1.95e-3$ & $1.17e-3$                                          & $\mathbf{7.38e-4}$                                \\ \cline{2-2}
\multicolumn{1}{c|}{}                                                                         & \multicolumn{1}{c|}{$6$~dB} & $9.17e-4$ & \multicolumn{1}{c|}{$6.20e-4$}                                         & $6.32e-4$ & \multicolumn{1}{c|}{$3.32e-4$}                                          & $2.99e-4$ & $1.72e-4$                                          & $\mathbf{7.65e-5}$                                \\ \hline
\multicolumn{1}{c|}{\multirow{3}{*}{\begin{tabular}[c]{@{}c@{}}BCH\\ $(63,30)$\end{tabular}}}   & \multicolumn{1}{c|}{$4$~dB} & $4.32e-2$ & \multicolumn{1}{c|}{$3.23e-2$}                                         & $4.07e-2$  & \multicolumn{1}{c|}{$2.49e-2$}                                          & $3.63e-2$ & $2.15e-2$                                          & $\mathbf{1.81e-2}$                                \\ \cline{2-2}
\multicolumn{1}{c|}{}                                                                         & \multicolumn{1}{c|}{$5$~dB} & $1.88e-2$ & \multicolumn{1}{c|}{$1.28e-2$}                                         & $1.72e-2$  & \multicolumn{1}{c|}{$8.60e-3$}                                          & $1.41e-2$ & $6.99e-3$                                          & $\mathbf{4.80e-3}$                                \\ \cline{2-2}
\multicolumn{1}{c|}{}                                                                         & \multicolumn{1}{c|}{$6$~dB} & $5.70e-3$ & \multicolumn{1}{c|}{$3.72e-3$}                                         & $4.96e-3$  & \multicolumn{1}{c|}{$2.11e-3$}                                          & $3.53e-3$ & $1.45e-3$                                          & $\mathbf{8.61e-4}$                                \\ \hline
\multicolumn{1}{c|}{\multirow{3}{*}{\begin{tabular}[c]{@{}c@{}}BCH\\ $(63,45)$\end{tabular}}}   & \multicolumn{1}{c|}{$4$~dB} & $1.58e-2$ & \multicolumn{1}{c|}{$1.16e-2$}                                         & $1.27e-2$  & \multicolumn{1}{c|}{$9.63e-3$}                                          & $1.14e-2$ & $8.34e-3$                                          & $\mathbf{6.41e-3}$                                \\ \cline{2-2}
\multicolumn{1}{c|}{}                                                                         & \multicolumn{1}{c|}{$5$~dB} & $4.41e-3$ & \multicolumn{1}{c|}{$2.87e-3$}                                         & $3.16e-3$  & \multicolumn{1}{c|}{$2.21e-3$}                                          & $2.64e-3$ & $1.72e-3$                                          & $\mathbf{1.05e-3}$                                \\ \cline{2-2}
\multicolumn{1}{c|}{}                                                                         & \multicolumn{1}{c|}{$6$~dB} & $7.57e-4$ & \multicolumn{1}{c|}{$4.13e-4$}                                         & $4.35e-4$  & \multicolumn{1}{c|}{$2.97e-4$}                                          & $3.38e-4$ & $2.00e-4$                                          & $\mathbf{1.02e-4}$                                \\ \hline
\multicolumn{1}{c|}{\multirow{3}{*}{\begin{tabular}[c]{@{}c@{}}Polar\\ $(64,22)$\end{tabular}}} & \multicolumn{1}{c|}{$4$~dB} & $4.03e-2$ & \multicolumn{1}{c|}{$3.50e-2$}                                         & $2.76e-2$  & \multicolumn{1}{c|}{$1.74e-2$}                                          & $1.44e-2$ & $1.11e-2$                                          & $\mathbf{8.59e-3}$                                \\ \cline{2-2}
\multicolumn{1}{c|}{}                                                                         & \multicolumn{1}{c|}{$5$~dB} & $1.66e-2$ & \multicolumn{1}{c|}{$1.33e-2$}                                         & $9.34e-3$  & \multicolumn{1}{c|}{$4.51e-3$}                                          & $3.91e-3$ & $2.30e-3$                                          & $\mathbf{1.66e-3}$                                \\ \cline{2-2}
\multicolumn{1}{c|}{}                                                                         & \multicolumn{1}{c|}{$6$~dB} & $5.17e-3$ & \multicolumn{1}{c|}{$3.42e-3$}                                         & $2.34e-3$  & \multicolumn{1}{c|}{$7.05e-4$}                                          & $1.08e-3$ & $2.73e-4$                                          & $\mathbf{1.52e-4}$                                \\ \hline
\multicolumn{1}{c|}{\multirow{3}{*}{\begin{tabular}[c]{@{}c@{}}Polar\\ $(64,32)$\end{tabular}}} & \multicolumn{1}{c|}{$4$~dB} & $1.74e-2$ & \multicolumn{1}{c|}{$1.29e-2$}                                         & $1.20e-2$  & \multicolumn{1}{c|}{$1.01e-2$}                                          & $9.04e-3$ & $7.84e-3$                                          & $\mathbf{5.62e-3}$                                \\ \cline{2-2}
\multicolumn{1}{c|}{}                                                                         & \multicolumn{1}{c|}{$5$~dB} & $5.84e-3$ & \multicolumn{1}{c|}{$3.99e-3$}                                         & $3.33e-3$  & \multicolumn{1}{c|}{$2.79e-3$}                                          & $2.39e-3$ & $1.97e-3$                                          & $\mathbf{1.23e-3}$                                \\ \cline{2-2}
\multicolumn{1}{c|}{}                                                                         & \multicolumn{1}{c|}{$6$~dB} & $1.39e-3$ & \multicolumn{1}{c|}{$8.63e-4$}                                         & $6.22e-4$  & \multicolumn{1}{c|}{$5.17e-4$}                                          & $4.35e-4$ & $3.48e-4$                                          & $\mathbf{1.60e-4}$                                \\ \hline
\multicolumn{1}{c|}{\multirow{3}{*}{\begin{tabular}[c]{@{}c@{}}Polar\\ $(64,48)$\end{tabular}}} & \multicolumn{1}{c|}{$4$~dB} & $6.75e-3$ & \multicolumn{1}{c|}{$7.44e-3$}                                         & $5.37e-3$  & \multicolumn{1}{c|}{$1.69e-1$}                                          & $4.17e-3$ & $4.77e-3$                                          & $\mathbf{3.43e-3}$                                \\ \cline{2-2}
\multicolumn{1}{c|}{}                                                                         & \multicolumn{1}{c|}{$5$~dB} & $1.42e-3$ & \multicolumn{1}{c|}{$1.62e-3$}                                         & $1.02e-3$  & \multicolumn{1}{c|}{$1.13e-3$}                                          & $6.95e-4$ & $8.04e-4$                                          & $\mathbf{4.95e-4}$                                \\ \cline{2-2}
\multicolumn{1}{c|}{}                                                                         & \multicolumn{1}{c|}{$6$~dB} & $1.83e-4$ & \multicolumn{1}{c|}{$2.15e-4$}                                         & $1.24e-4$  & \multicolumn{1}{c|}{$1.42e-4$}                                          & $7.47e-5$ & $8.58e-5$                                          & $\mathbf{4.27e-5}$                                \\ \hline\hline
\end{tabular}
\caption{A comparison of BER for three different methods at three different SNR values~($4$~dB, $5$~dB, $6$~dB).
This table shows the BER performance for $N=2$ and $d=\{32,64,128\}$.
Best results are in \textbf{bold}.
\label{tab_n2}}
\end{table*}

\begin{figure*}[!t]
\centering
\begin{subfigure}[b]{0.246\textwidth}
    \centering
    \includegraphics[width=\textwidth]{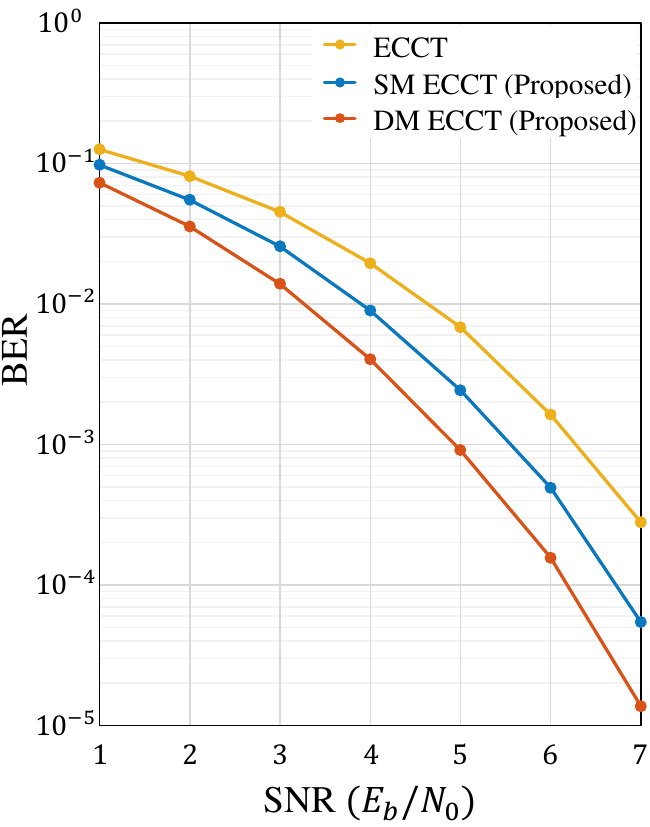} % Reduce the figure size so that it is slightly narrower than the column. Don't use precise values for figure width.This setup will avoid overfull boxes.
\caption{BCH~$(31,11)$}
\end{subfigure}
\hfill
\begin{subfigure}[b]{0.246\textwidth}
    \centering
    \includegraphics[width=\textwidth]{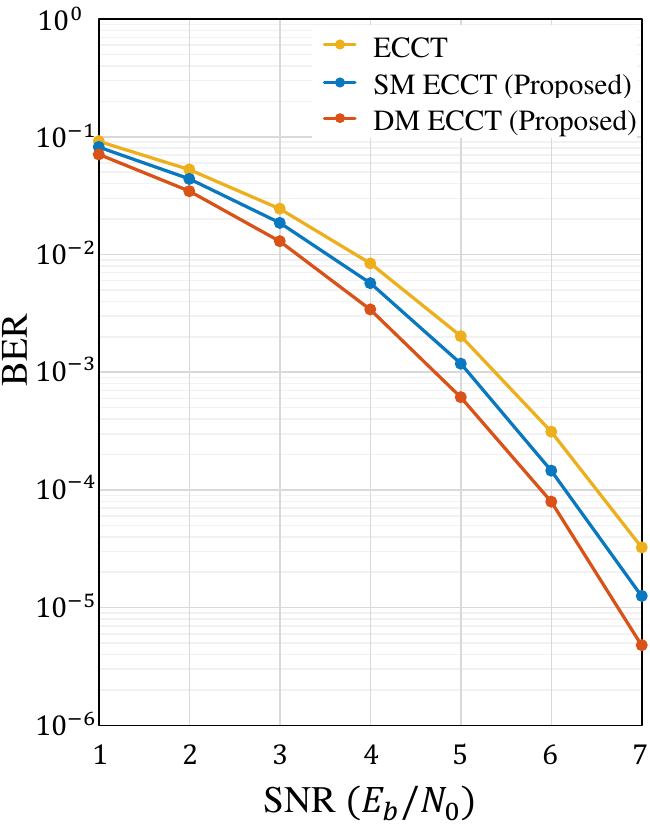} % Reduce the figure size so that it is slightly narrower than the column. Don't use precise values for figure width.This setup will avoid overfull boxes.
\caption{BCH~$(31,16)$}
\end{subfigure}
\hfill
\begin{subfigure}[b]{0.246\textwidth}
    \centering
    \includegraphics[width=\textwidth]{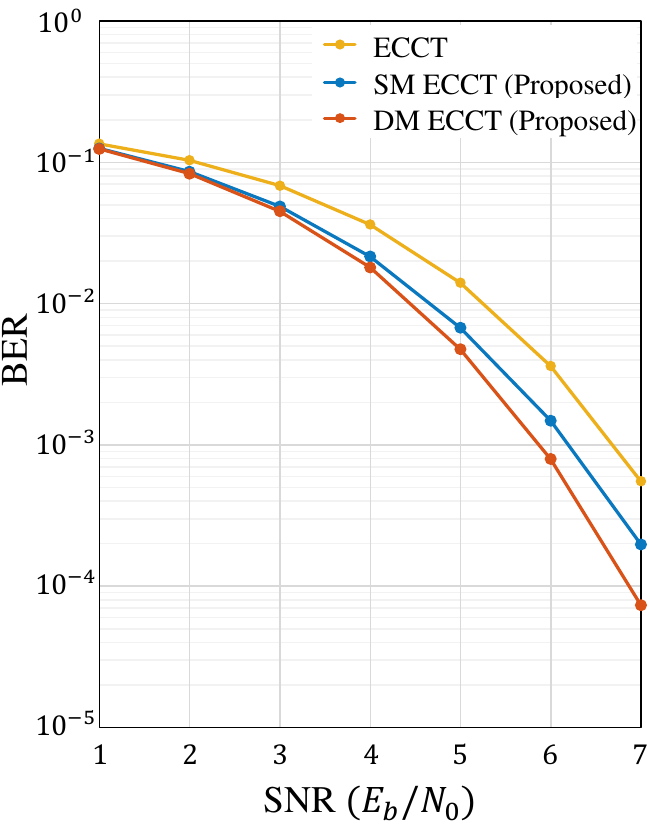} % Reduce the figure size so that it is slightly narrower than the column. Don't use precise values for figure width.This setup will avoid overfull boxes.
\caption{BCH~$(63,30)$}
\end{subfigure}
\hfill
\begin{subfigure}[b]{0.246\textwidth}
    \centering
    \includegraphics[width=\textwidth]{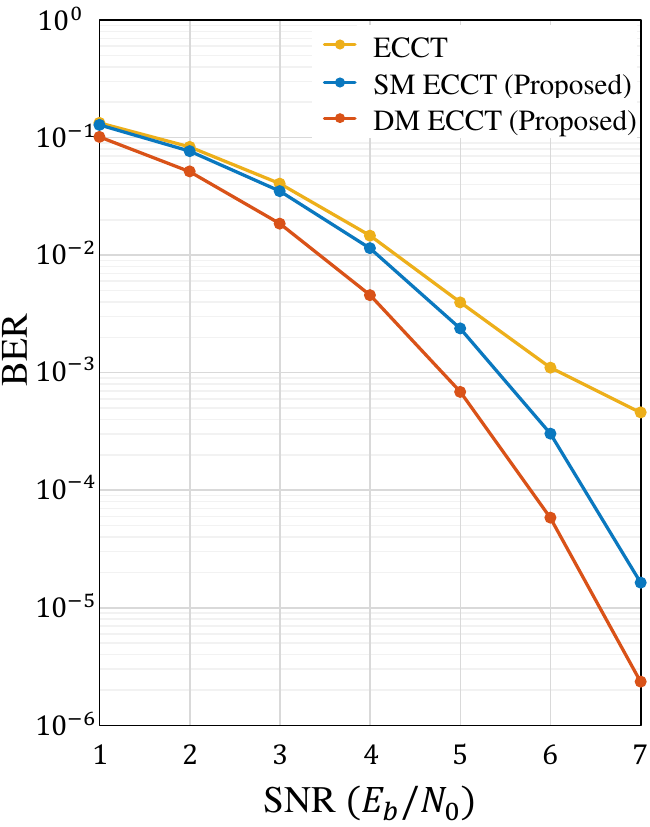} % Reduce the figure size so that it is slightly narrower than the column. Don't use precise values for figure width.This setup will avoid overfull boxes.
\caption{Polar~$(64,22)$}
\end{subfigure}
\caption{The BER performance of BCH and polar codes for $N=2$ and $d=128$.
\label{fig_N2}}
\end{figure*}

\begin{table*}[!t]
\centering
\begin{tabular}{ccccccccc}
\hline\hline
\multicolumn{9}{c}{$N=6$}                                                                                                                                                                                                                                                                                           \\ \hline\hline
\multicolumn{2}{c|}{$d$}                                                                                                    & \multicolumn{2}{c|}{32}                     & \multicolumn{2}{c|}{64}                     & \multicolumn{3}{c}{128}                                                                   \\ \hline
\multicolumn{1}{c|}{Code}                                                                     & \multicolumn{1}{l|}{SNR}  & ECCT       & \multicolumn{1}{c|}{\begin{tabular}[c]{@{}c@{}}SM\\ ECCT\end{tabular}} & ECCT       & \multicolumn{1}{c|}{\begin{tabular}[c]{@{}c@{}}SM\\ ECCT\end{tabular}} & ECCT       & \begin{tabular}[c]{@{}c@{}}SM\\ ECCT\end{tabular}         & \begin{tabular}[c]{@{}c@{}}DM\\ ECCT\end{tabular} \\ \hline\hline
\multicolumn{1}{c|}{\multirow{3}{*}{\begin{tabular}[c]{@{}c@{}}BCH\\ $(31,11)$\end{tabular}}}   & \multicolumn{1}{c|}{$4$~dB} & $1.22e-2$ & \multicolumn{1}{c|}{$5.22e-3$}  & $8.35e-3$ & \multicolumn{1}{c|}{$2.48e-3$}  & $4.37e-3$ & $1.72e-3$          & $\mathbf{8.70e-4}$                                       \\ \cline{2-2}
\multicolumn{1}{c|}{}                                                                         & \multicolumn{1}{c|}{$5$~dB} & $3.28e-3$ & \multicolumn{1}{c|}{$1.27e-3$}  & $1.92e-3$ & \multicolumn{1}{c|}{$5.91e-4$}  & $9.69e-4$ & $3.22e-4$          & $\mathbf{8.57e-5}$                                       \\ \cline{2-2}
\multicolumn{1}{c|}{}                                                                         & \multicolumn{1}{c|}{$6$~dB} & $4.94e-4$ & \multicolumn{1}{c|}{$2.10e-4$}  & $2.37e-4$ & \multicolumn{1}{c|}{$8.23e-5$}  & $1.10e-4$ & $4.89e-5$          & $\mathbf{6.70e-6}$                                       \\ \hline
\multicolumn{1}{c|}{\multirow{3}{*}{\begin{tabular}[c]{@{}c@{}}BCH\\ $(31,16)$\end{tabular}}}   & \multicolumn{1}{c|}{$4$~dB} & $5.55e-3$ & \multicolumn{1}{c|}{$2.79e-3$}  & $3.50e-3$ & \multicolumn{1}{c|}{$3.11e-3$}  & $2.97e-3$ & $1.71e-3$          & $\mathbf{1.07e-3}$                                       \\ \cline{2-2}
\multicolumn{1}{c|}{}                                                                         & \multicolumn{1}{c|}{$5$~dB} & $1.02e-3$ & \multicolumn{1}{c|}{$4.43e-4$}  & $6.50e-4$ & \multicolumn{1}{c|}{$5.24e-4$}  & $4.66e-4$ & $2.52e-4$          & $\mathbf{8.35e-5} $                                      \\ \cline{2-2}
\multicolumn{1}{c|}{}                                                                         & \multicolumn{1}{c|}{$6$~dB} & $1.12e-4$ & \multicolumn{1}{c|}{$3.11e-5$}  & $6.65e-5$ & \multicolumn{1}{c|}{$5.66e-5$}  & $4.76e-5$ & $2.35e-5$          & $\mathbf{5.75e-6} $                                      \\ \hline
\multicolumn{1}{c|}{\multirow{3}{*}{\begin{tabular}[c]{@{}c@{}}BCH\\ $(63,30)$\end{tabular}}}   & \multicolumn{1}{c|}{$4$~dB} & $8.41e-3$ & \multicolumn{1}{c|}{$5.63e-3$}  & $7.36e-3$ & \multicolumn{1}{c|}{$4.41e-3$}  & $1.87e-2$ &$ {9.00e-3}$ & $\mathbf{8.04e-3}$                                                \\ \cline{2-2}
\multicolumn{1}{c|}{}                                                                         & \multicolumn{1}{c|}{$5$~dB} & $1.37e-3$ & \multicolumn{1}{c|}{$8.33e-4$}  & $1.06e-3$ & \multicolumn{1}{c|}{$5.24e-4$}  & $4.41e-3$ & ${1.96e-3}$ & $\mathbf{1.18e-3}$                                                \\ \cline{2-2}
\multicolumn{1}{c|}{}                                                                         & \multicolumn{1}{c|}{$6$~dB} & $1.10e-4$ & \multicolumn{1}{c|}{$5.09e-5$}  & $7.02e-5$ & \multicolumn{1}{c|}{$2.68e-5$}  & $5.28e-4$ & ${2.51e-4}$ & $\mathbf{7.61e-5}$                                                \\ \hline
\multicolumn{1}{c|}{\multirow{3}{*}{\begin{tabular}[c]{@{}c@{}}BCH\\ $(63,45)$\end{tabular}}}   & \multicolumn{1}{c|}{$4$~dB} & $1.00e-2$ & \multicolumn{1}{c|}{$5.03e-3$}  & $6.12e-3$ & \multicolumn{1}{c|}{$2.05e-3$}  & $4.53e-3$ & $3.76e-3$          & $\mathbf{2.74e-3}$                                       \\ \cline{2-2}
\multicolumn{1}{c|}{}                                                                         & \multicolumn{1}{c|}{$5$~dB} & $2.49e-3$ & \multicolumn{1}{c|}{$8.95e-4$}  & $1.59e-3$ & \multicolumn{1}{c|}{$2.49e-4$}  & $5.58e-4$ & $4.32e-4$          & $\mathbf{2.62e-4}$                                       \\ \cline{2-2}
\multicolumn{1}{c|}{}                                                                         & \multicolumn{1}{c|}{$6$~dB} & $6.83e-4$ & \multicolumn{1}{c|}{$1.07e-4$}  & $5.42e-4$ & \multicolumn{1}{c|}{$1.82e-5$}  & $3.15e-5$ & $1.61e-5$          & $\mathbf{9.31e-6}$                                       \\ \hline
\multicolumn{1}{c|}{\multirow{3}{*}{\begin{tabular}[c]{@{}c@{}}Polar\\ $(64,22)$\end{tabular}}} & \multicolumn{1}{c|}{$4$~dB} & $5.22e-3$ & \multicolumn{1}{c|}{$3.32e-3$}  & $2.52e-3$ & \multicolumn{1}{c|}{$1.59e-3$}  & $2.16e-3$ & $5.84e-4$          & $\mathbf{4.82e-4}$                                       \\ \cline{2-2}
\multicolumn{1}{c|}{}                                                                         & \multicolumn{1}{c|}{$5$~dB} & $9.64e-4$ & \multicolumn{1}{c|}{$5.01e-4$}  & $3.50e-4$ & \multicolumn{1}{c|}{$1.76e-4$}  & $1.76e-4$ & $3.28e-5$          & $\mathbf{2.94e-5}$                                       \\ \cline{2-2}
\multicolumn{1}{c|}{}                                                                         & \multicolumn{1}{c|}{$6$~dB} & $1.01e-4$ & \multicolumn{1}{c|}{$4.44e-5$}  & $2.66e-5$ & \multicolumn{1}{c|}{$1.39e-5$}  & $1.22e-5$ & $1.37e-6$          & $\mathbf{6.39e-7}$                                       \\ \hline
\multicolumn{1}{c|}{\multirow{3}{*}{\begin{tabular}[c]{@{}c@{}}Polar\\ $(64,32)$\end{tabular}}} & \multicolumn{1}{c|}{$4$~dB} & $3.59e-3$ & \multicolumn{1}{c|}{$3.06e-3$}  & $2.30e-3$ & \multicolumn{1}{c|}{$2.09e-3$}  & $1.25e-3$ & $9.10e-4$          & $\mathbf{9.03e-4}$                                       \\ \cline{2-2}
\multicolumn{1}{c|}{}                                                                         & \multicolumn{1}{c|}{$5$~dB} & $5.21e-4$ & \multicolumn{1}{c|}{$4.22e-4$}  & $2.94e-4$ & \multicolumn{1}{c|}{$2.66e-4$}  & $1.25e-4$ & $8.28e-5$          & $\mathbf{8.07e-5}$                                       \\ \cline{2-2}
\multicolumn{1}{c|}{}                                                                         & \multicolumn{1}{c|}{$6$~dB} & $4.56e-5$ & \multicolumn{1}{c|}{$3.10e-5$}  & $1.98e-5$ & \multicolumn{1}{c|}{$1.46e-5$}  & $6.57e-6$ & $4.47e-6$          & $\mathbf{3.70e-6}$                                       \\ \hline
\multicolumn{1}{c|}{\multirow{3}{*}{\begin{tabular}[c]{@{}c@{}}Polar\\ $(64,48)$\end{tabular}}} & \multicolumn{1}{c|}{$4$~dB} & $6.75e-3$ & \multicolumn{1}{c|}{$7.44e-3$}  & $5.37e-3$  & \multicolumn{1}{c|}{$1.69e-1$}   & $2.05e-3$ & $\mathbf{1.68e-3}$          & ${1.70e-3}$                                       \\ \cline{2-2}
\multicolumn{1}{c|}{}                                                                         & \multicolumn{1}{c|}{$5$~dB} & $1.42e-3$ & \multicolumn{1}{c|}{$1.62e-3$}  & $1.02e-3$  & \multicolumn{1}{c|}{$1.13e-3$}   & $2.50e-4$ & $2.12e-4$          & $\mathbf{2.11e-4}$                                       \\ \cline{2-2}
\multicolumn{1}{c|}{}                                                                         & \multicolumn{1}{c|}{$6$~dB} & $1.83e-4$ & \multicolumn{1}{c|}{$2.15e-4$}  & $1.24e-4$  & \multicolumn{1}{c|}{$1.42e-4$}   & $1.77e-5$ & $1.83e-5$          & $\mathbf{1.73e-5}$                                       \\ \hline\hline
\end{tabular}
\caption{A comparison of BER for three different methods at three different SNR values~($4$~dB, $5$~dB, $6$~dB).
This table shows the BER performance for $N=6$ and $d=\{32,64,128\}$.
Best results are in \textbf{bold}.
\label{tab_n6}}
\end{table*}

\begin{figure*}[!t]
\centering
\begin{subfigure}[b]{0.246\textwidth}
    \centering
    \includegraphics[width=\textwidth]{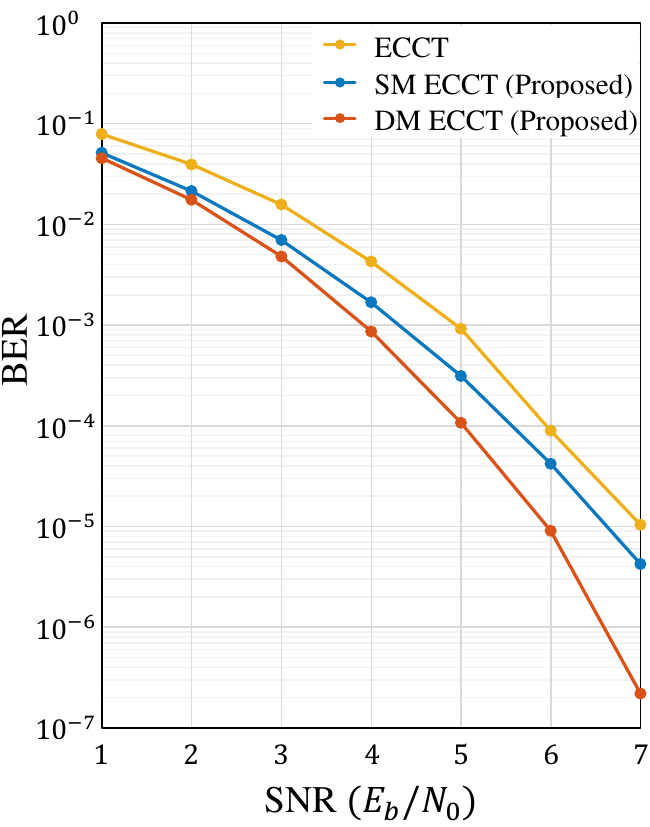} % Reduce the figure size so that it is slightly narrower than the column. Don't use precise values for figure width.This setup will avoid overfull boxes.
\caption{BCH~$(31,11)$}
\end{subfigure}
\hfill
\begin{subfigure}[b]{0.246\textwidth}
    \centering
    \includegraphics[width=\textwidth]{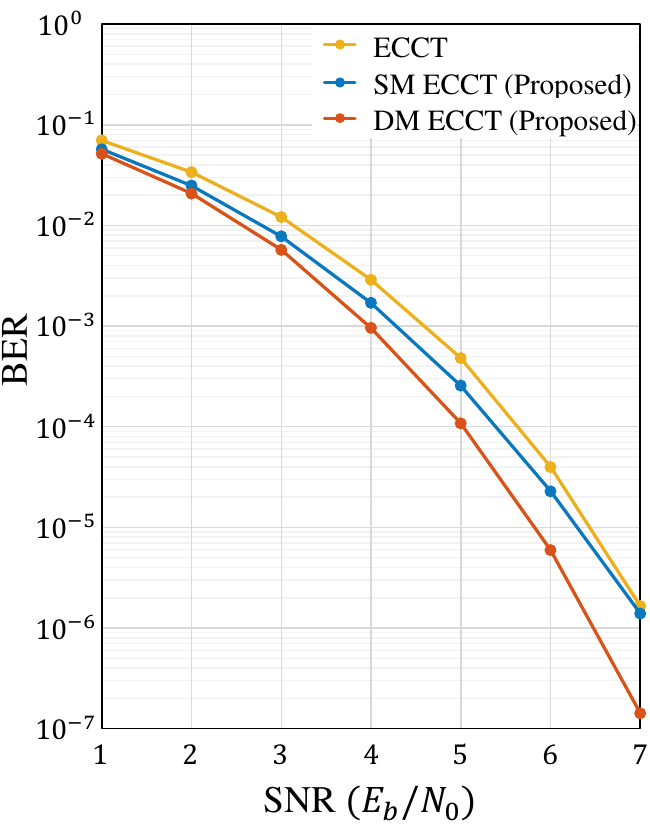} % Reduce the figure size so that it is slightly narrower than the column. Don't use precise values for figure width.This setup will avoid overfull boxes.
\caption{BCH~$(31,16)$}
\end{subfigure}
\hfill
\begin{subfigure}[b]{0.246\textwidth}
    \centering
    \includegraphics[width=\textwidth]{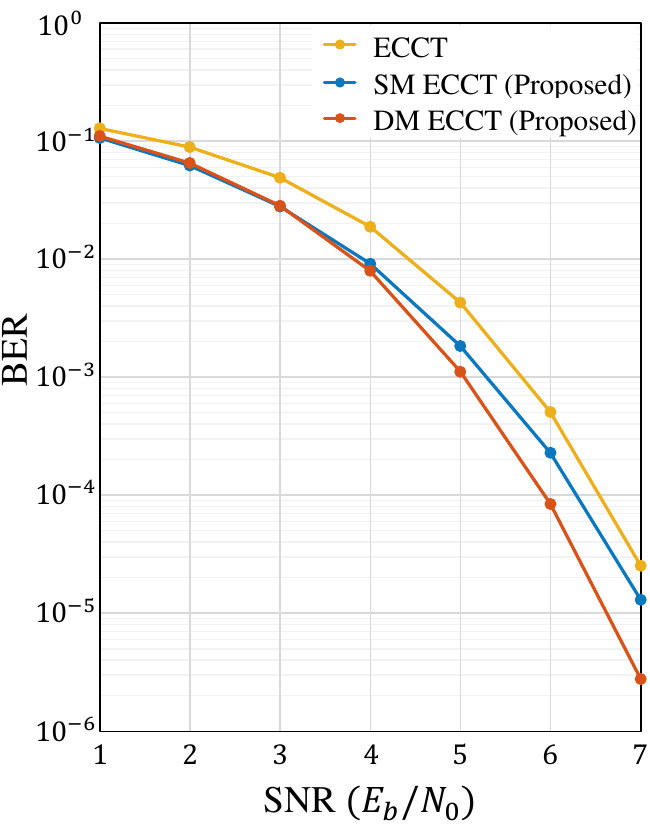} % Reduce the figure size so that it is slightly narrower than the column. Don't use precise values for figure width.This setup will avoid overfull boxes.
\caption{BCH~$(63,30)$}
\end{subfigure}
\hfill
\begin{subfigure}[b]{0.246\textwidth}
    \centering
    \includegraphics[width=\textwidth]{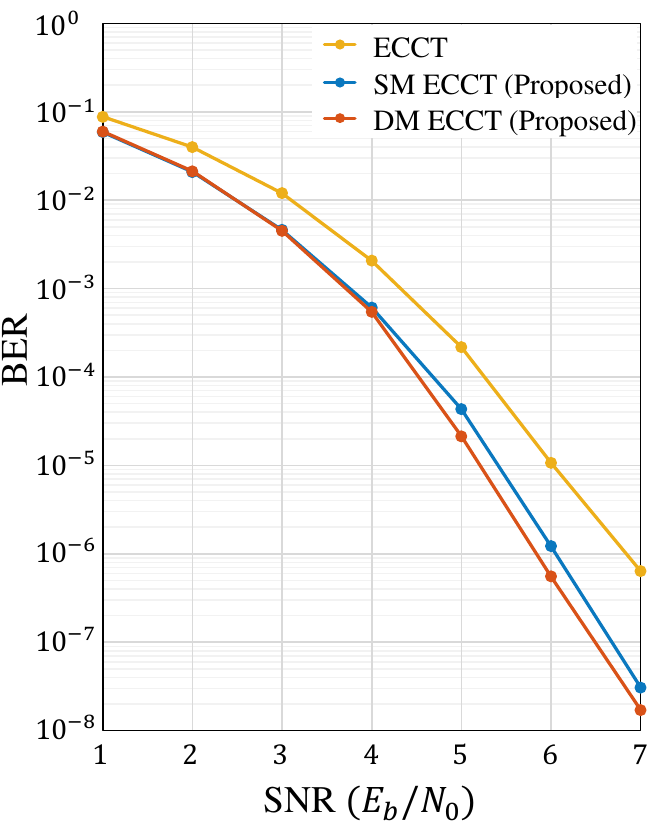} % Reduce the figure size so that it is slightly narrower than the column. Don't use precise values for figure width.This setup will avoid overfull boxes.
\caption{Polar~$(64,22)$}
\end{subfigure}
\caption{The BER performance of BCH and polar codes for $N=6$ and $d=128$.
\label{fig_N6}}
\end{figure*}

\section{Training}

The goal of the proposed decoder is to learn the multiplicative noise $\tilde{z}_s$ in (\ref{equ_multi_noise}) and recover an original transmitted signal $x$.
We can obtain the multiplicative noise by $\tilde{z}_s = \tilde{z}_s x^2_s = yx_s$.
Then, the target multiplicative noise for binary cross-entropy loss function is defined by $\tilde{z} =\text{bin}(\text{sign}(yx_s))$.
Finally, the cross-entropy loss function for a received codeword $y$ is defined as
\begin{equation*}
    \mathcal{L} = -\sum^{n}_{i=1}\tilde{z}_i \log(\sigma(f(y)))+(1-\tilde{z}_i)\log(1-\sigma(f(y))).
\end{equation*}

To compare with the conventional ECCT fairly, we adopt the same training setup as used in the previous work.
We use the Adam optimizer~\cite{b_adam} and conduct 1000 epochs.
Each epoch consists of 1000 minibatches, where each minibatch is composed of 128 samples.

\begin{figure}[!h]
\centering
\begin{subfigure}[b]{\columnwidth}
    \centering
    \includegraphics[width=\columnwidth]{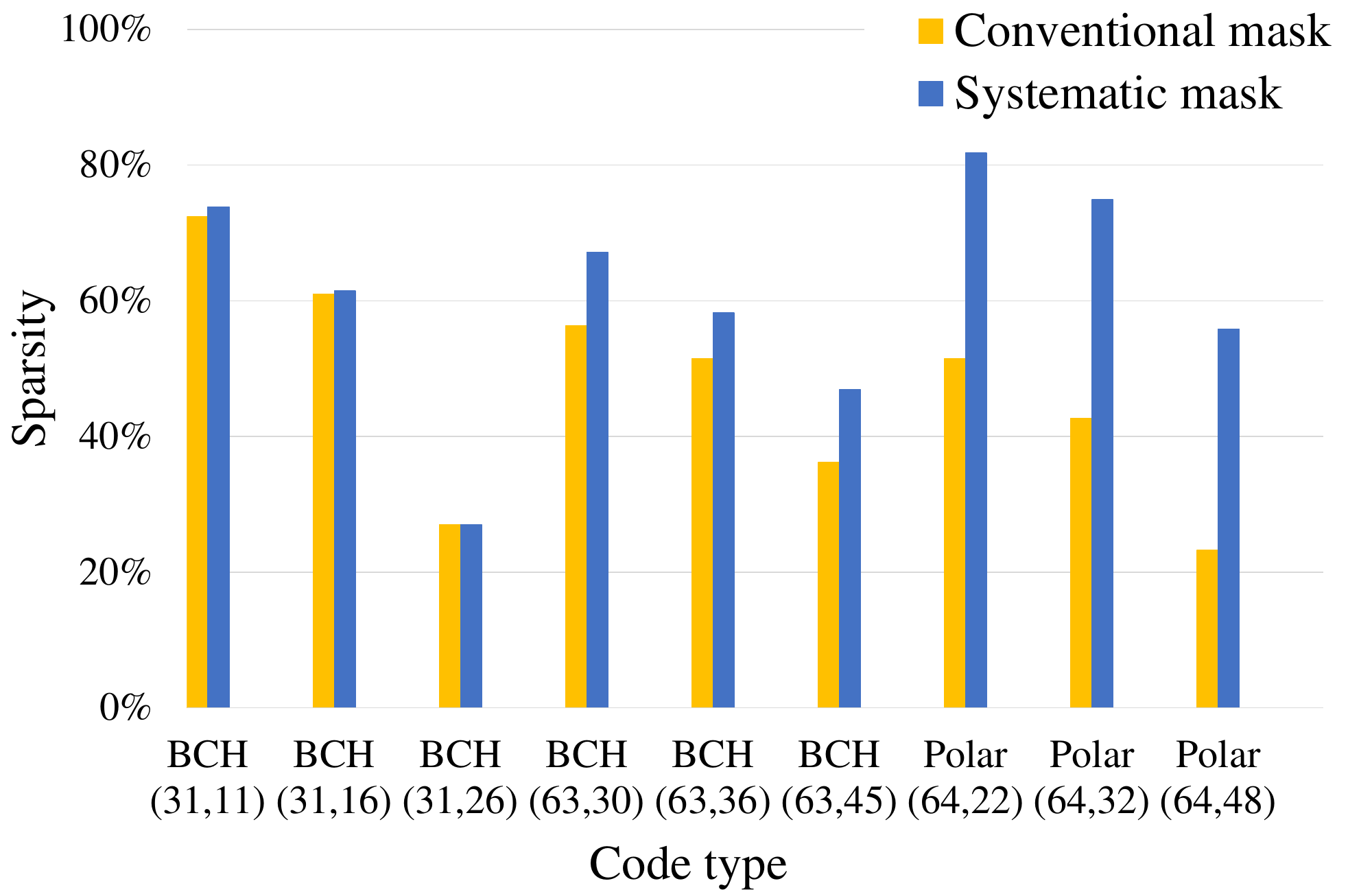} % Reduce the figure size so that it is slightly narrower than the column. Don't use precise values for figure width.This setup will avoid overfull boxes.
\caption{Sparsity \label{subfig_sparsity}}
\end{subfigure}
\begin{subfigure}[b]{\columnwidth}
    \centering
    \includegraphics[width=\columnwidth]{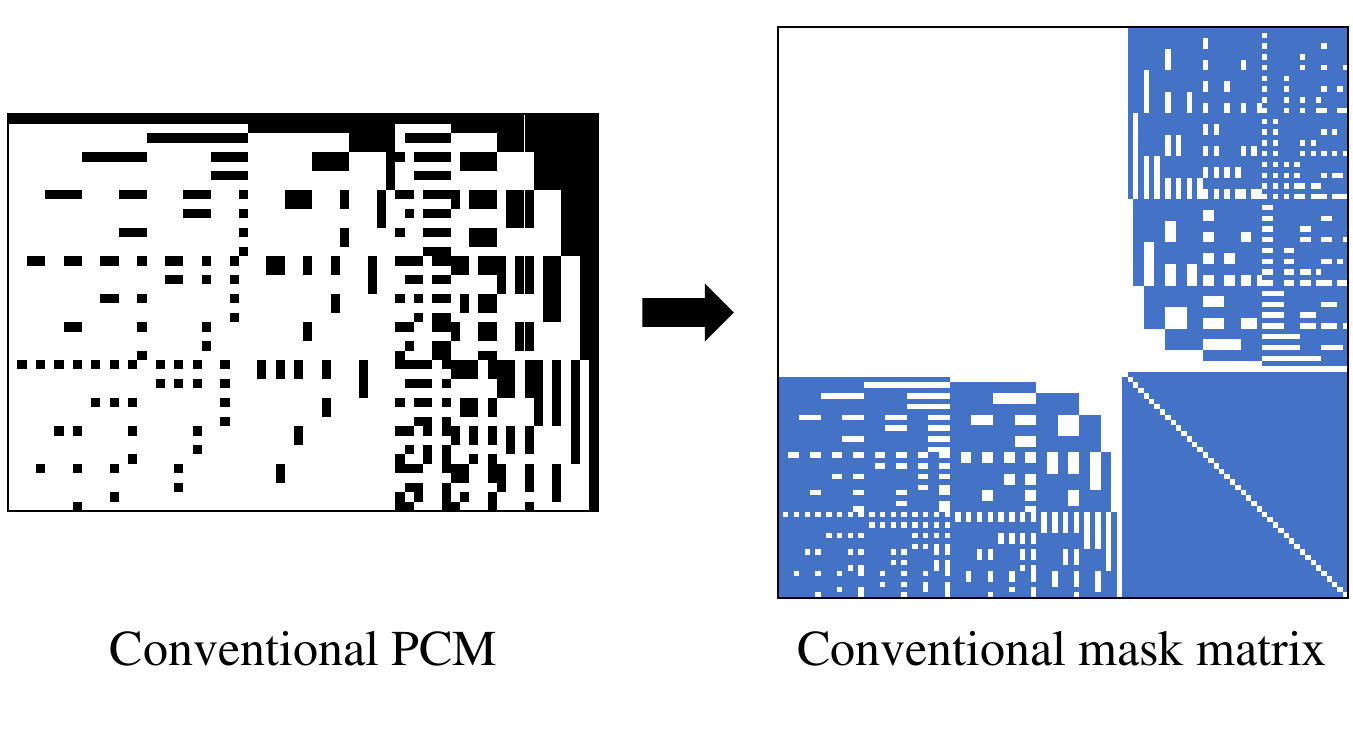}
\caption{Conventional ECCT}
\end{subfigure}
\hfill
\begin{subfigure}[b]{\columnwidth}
    \centering
    \includegraphics[width=\columnwidth]{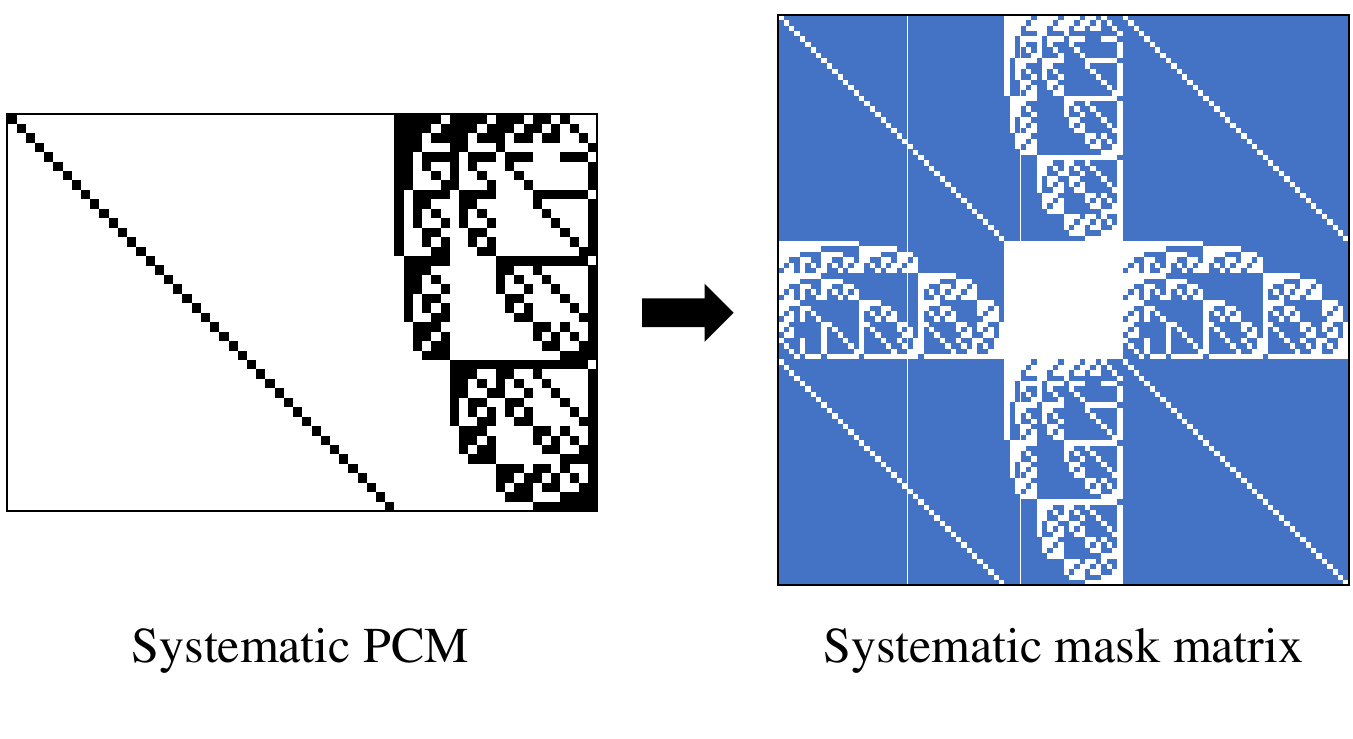}
\caption{Proposed ECCT}
\end{subfigure}

\caption{(a) Sparsity comparison of the self-attention map employing non-systematic and systematic mask matrices for different codes. (b) The PCM and mask matrix of polar code $(64,22)$, which is utilized in the conventional ECCT. (c) The PCM and mask matrix of polar code $(64,22)$, which is utilized in our work.
\label{fig_polar_mask}}
\end{figure}

\section{Experiments}

To evaluate the efficacy of our proposed methods, we assess the bit error rate~(BER) performance of both SM ECCT and DM ECCT for BCH codes and polar codes, and then compare with the conventional ECCT~\cite{b_ECCT}.
The implementation of the conventional ECCT is taken from \cite{b_ECCT_code}.
For the testing, we collect at least 500 frame errors at each signal-to-noise ratio~(SNR) value for at least $10^{5}$ random codewords.

{The DM ECCT utilizes the systematic and conventional mask matrices for two self-attention blocks.
For polar codes, however, we utilize the systematic mask matrix and the mask matrix constructed based on the modified conventional PCM.
In the following section, it will be mentioned that the upper $n\times n$ submatrix in the conventional mask matrix of polar code is always unmasked and the DM ECCT obtains the dense self-attention map.
The dense self-attention map hinders the ECCT from effectively learning the relationship between codeword bits.
Thus, we modify the conventional PCM of the polar codes by the row reduction.
Starting from the first row, when the row contains all positions of one's of the next row, we eliminate the ones in the row.
Utilizing both systematic and modified conventional mask matrices provides diversity gains to the DM ECCT.}

Tables~\ref{tab_n2} and \ref{tab_n6} show the BER performances of the conventional ECCT, the proposed SM ECCT, and the proposed DM ECCT for $N=\{2,6\}$ and \mbox{$d=\{32,64,128\}$}.
The first, second, and the third rows of each codes correspond to the BER performances at SNR $4$~dB, $5$~dB, and $6$~dB, respectively.
For the same $N$ and SNR, the proposed DM ECCT achieves the best BER in most cases, except for BCH code $(63,30)$ when $N=6$ and $d=128$.
The SM ECCT outperforms the conventional ECCT despite its lower computational complexity.
Figures~\ref{fig_N2} and \ref{fig_N6} show the decoding performance of BCH codes and polar codes for $N=2$, $d=128$ and $N=6$, $d=128$, respectively.
The DM ECCT, notably at a low code rate, significantly outperforms the conventional ECCT by more than 1 dB for high SNR.
Furthermore, the decoding performance for both SM ECCT and DM ECCT continues to decrease, even when the ECCT curve starts to flatten, indicating the potential of the proposed ECCTs to perform well at high SNR points.
Given that the conventional ECCT outperforms previous neural network-based decoding algorithms as noted by Choukroun and Wolf~\cite{b_ECCT}, our work can be recognized as the new state-of-the art solution for ECC decoding.

\section{Discussion}

\subsection{Complexity Analysis}

Figure~\ref{fig_polar_mask}(a) shows the sparsity of the self-attention map employing conventional and systematic masks with respect to the full self-attention map without masking.
Utilizing the systematic mask, sparsity levels rise from 72\% to 74\% for BCH code $(31, 11)$, from 56\% to 67\% for BCH code $(63, 30)$, and dramatically from 52\% to 82\% for polar code $(64, 22)$.
As depicted in Figures~\ref{fig_polar_mask}(b) and \ref{fig_polar_mask}(c), the systematic mask matrix shows a notably larger portion of masking positions than the conventional mask matrix, especially for polar codes.
This is attributed to the fact that the first row in the conventional PCM of the polar codes is the all-ones vector.
In such case, the upper $n\times n$ submatrix of the conventional mask matrix is always unmasked according to the Algorithm~\ref{alg_mask}.
However, the first row of the systematic PCM is not the all-ones vector and a large portion of the upper $n\times n$ submatrix is masked.
This sparsity improvement in the self-attention maps leads to a reduction in computational complexity.

In terms of the computational complexity, the use of the proposed systematic mask matrix contributes to reducing the computational complexity.
For the DM ECCT, it requires twice of a complexity in the decoder layer, as it has two input embedded vectors that pass through all blocks in the decoder layer.
However, the architecture of DM ECCT facilitates the parallel operations in the decoder layers; hence, DM ECCT can improve decoding performance while maintaining the decoding latency.

\section{Conclusion}

In this paper, we aimed to improve the performance of ECC decoding using a novel architecture of the ECCT.
We first proposed the systematic mask matrix, which is more suitable for the ECCT than the conventional mask matrix.
Additionally, we proposed the novel architecture of DM ECCT by employing two mutually complementary mask matrices.

Through extensive simulations, we demonstrated that the proposed methods outperform the conventional ECCT.
We first achieved performance and computational complexity enhancements by the systematic mask matrix.
The sparsity incurred by the systematic mask prompts the ECCT to focus on more important positions, enhancing decoding performance.
In particular, more pronounced improvements are observed in low-rate codes.
Traditionally, the systematic form of the matrix has been employed for efficient encoding in conventional decoders (e.g., BP, MS decoders).
However, our results highlight its critical importance in the decoding process of ECCT as well.

Also, we showed that the proposed DM ECCT architecture contributes to decoding performance improvement.
We utilized two different mask matrices, systematic and conventional masks, in a parallel manner, which provided diversity gains to the decoder.
The DM ECCT notably enhances decoding performance over the conventional ECCT for both BCH codes and polar codes, achieving the state-of-the-art decoding performance among neural network-based decoders with considerable margins.

\bibliography{arxiv24}

\begin{thebibliography}{19}
\providecommand{\natexlab}[1]{#1}

\bibitem[{Bennatan, Choukroun, and Kisilev(2018)}]{b_preproc}
Bennatan, A.; Choukroun, Y.; and Kisilev, P. 2018.
\newblock {Deep learning for decoding of linear codes-a syndrome-based
  approach}.
\newblock In \emph{Proceedings of 2018 IEEE International Symposium on
  Information Theory (ISIT)}, 1595--1599. {IEEE}.

\bibitem[{Buchberger et~al.(2021)Buchberger, Hager, Pfister, Schmalen, and
  Amat}]{b_6}
Buchberger, A.; Hager, C.; Pfister, H.~D.; Schmalen, L.; and Amat, A. G.~I.
  2021.
\newblock Pruning and quantizing neural belief propagation decoders.
\newblock \emph{IEEE Journal of Selected Areas in Communications}, 39(7):
  1957--1966.

\bibitem[{Choukroun and Wolf(2022{\natexlab{a}})}]{b_ECCT}
Choukroun, Y.; and Wolf, L. 2022{\natexlab{a}}.
\newblock {Error correction code transformer}.
\newblock In \emph{Advances in Neural Information Processing Systems}.

\bibitem[{Choukroun and Wolf(2022{\natexlab{b}})}]{b_ECCT_code}
Choukroun, Y.; and Wolf, L. 2022{\natexlab{b}}.
\newblock Error correction code transformer.
\newblock \url{https://github.com/yoniLc/ECCT}.
\newblock Accessed: 2023-05-22.

\bibitem[{Dai et~al.(2021)Dai, Tan, Si, Niu, Chen, Poor, and Cui}]{b_2}
Dai, J.; Tan, K.; Si, Z.; Niu, K.; Chen, M.; Poor, H.~V.; and Cui, S. 2021.
\newblock Learning to decode protograph LDPC codes.
\newblock \emph{IEEE Journal of Selected Areas in Communications}, 39(7):
  1983--1999.

\bibitem[{Dorner et~al.(2018)Dorner, Cammerer, Hoydis, and Brink}]{b_comm4}
Dorner, S.; Cammerer, S.; Hoydis, J.; and Brink, S. 2018.
\newblock Deep learning based communication over the air.
\newblock \emph{IEEE Journal of Selected Topics in Signal Processing}, 12(1):
  132--143.

\bibitem[{Dosovitskiy et~al.(2021)Dosovitskiy, Beyer, Kolesnikov, Weissenborn,
  Zhai, Unterthiner, Dehghani, Minderer, Heigold, Gelly, Uszkoreit, and
  Houlsby}]{b_vit}
Dosovitskiy, A.; Beyer, L.; Kolesnikov, A.; Weissenborn, D.; Zhai, X.;
  Unterthiner, T.; Dehghani, M.; Minderer, M.; Heigold, G.; Gelly, S.;
  Uszkoreit, J.; and Houlsby, N. 2021.
\newblock {An image is worth $16\times16$ words: Transformers for image
  recognition at scale}.
\newblock In \emph{International Conference on Learning Representations
  (ICLR)}.

\bibitem[{Kim et~al.(2018)Kim, Jiang, Rana, Kannan, Oh, and
  Viswanath}]{b_comm1}
Kim, H.; Jiang, Y.; Rana, R.; Kannan, S.; Oh, S.; and Viswanath, P. 2018.
\newblock {Communication algorithms via deep learning}.
\newblock In \emph{International Conference on Learning Representations
  (ICLR)}.

\bibitem[{Kim, Oh, and Viswanath(2020)}]{b_comm2}
Kim, H.; Oh, S.; and Viswanath, P. 2020.
\newblock Physical layer communication via deep learning.
\newblock \emph{IEEE Journal of Selected Topics in Information Theory}, 1(1):
  5--18.

\bibitem[{Kingma and Ba(2014)}]{b_adam}
Kingma, D.~P.; and Ba, J. 2014.
\newblock {Adam: A method for stochastic optimization}.
\newblock In \emph{arXiv preprint arXiv:1412.6980}.

\bibitem[{Kwak et~al.(2022)Kwak, Kim, Kim, Kim, and No}]{b_5}
Kwak, H.-Y.; Kim, J.-W.; Kim, Y.; Kim, S.-H.; and No, J.-S. 2022.
\newblock Neural min-sum decoding for generalized LDPC codes.
\newblock \emph{IEEE Communications Letters}, 26(12): 2841--2845.

\bibitem[{Liu et~al.(2021)Liu, Lin, Cao, Hu, Wei, Zhang, Lin, and Guo}]{b_swin}
Liu, Z.; Lin, Y.; Cao, Y.; Hu, H.; Wei, Y.; Zhang, Z.; Lin, S.; and Guo, B.
  2021.
\newblock {Swin transformer: Hierarchical vision transformer using shifted
  windows}.
\newblock In \emph{IEEE/CVF International Conference on Computer Vision
  (ICCV)}.

\bibitem[{Lugosch and Gross(2017)}]{b_7}
Lugosch, L.; and Gross, W.~J. 2017.
\newblock {Neural offset min-sum decoding}.
\newblock In \emph{Proceedings of 2017 IEEE International Symposium on
  Information Theory (ISIT)}, 1316--1365. {IEEE}.

\bibitem[{Nachmani, Beery, and Burshtein(2016)}]{b_comm3}
Nachmani, E.; Beery, Y.; and Burshtein, D. 2016.
\newblock {Learning to decode linear codes using deep learning}.
\newblock In \emph{2016 54th Annual Allerton Conference on Communications,
  Control, and Computing (Allerton)}, 341--346. {IEEE}.

\bibitem[{Nachmani et~al.(2018)Nachmani, Marciano, Lugosch, Gross, Burshtein,
  and Beery}]{b_1}
Nachmani, E.; Marciano, E.; Lugosch, L.; Gross, W.~J.; Burshtein, D.; and
  Beery, Y. 2018.
\newblock Deep learning methods for improved decoding of linear codes.
\newblock \emph{IEEE Journal of Selected Topics in Signal Processing}, 12(1):
  119--131.

\bibitem[{Nachmani and Wolf(2019)}]{b_3}
Nachmani, E.; and Wolf, L. 2019.
\newblock {Hyper-graph-network decoders for block codes}.
\newblock In \emph{Advances in Neural Information Processing Systems},
  2326--2336.

\bibitem[{Nachmani and Wolf(2021)}]{b_4}
Nachmani, E.; and Wolf, L. 2021.
\newblock {Autoregressive belief propagation for decoding block codes}.
\newblock In \emph{arXiv preprint arXiv:2103.11780}.

\bibitem[{Richardson and Urbanke(2008)}]{b_modern}
Richardson, T.~J.; and Urbanke, R. 2008.
\newblock \emph{Modern coding theory}.
\newblock Cambridge Univ. Press.

\bibitem[{Vaswani et~al.(2017)Vaswani, Shazeer, Parmar, Uszkoreit, Jones,
  Gomez, Kaiser, and Polosukhin}]{b_transformer}
Vaswani, A.; Shazeer, N.; Parmar, N.; Uszkoreit, J.; Jones, L.; Gomez, A.~N.;
  Kaiser, {\L}.; and Polosukhin, I. 2017.
\newblock {Attention is all you need}.
\newblock In \emph{Advances in Neural Information Processing Systems},
  5998--6008.

\end{thebibliography}

\end{document}